\documentclass[conference,compsoc]{IEEEtran}
\ifCLASSOPTIONcompsoc
  \usepackage[nocompress]{cite}
\else
  \usepackage{cite}
\fi
\ifCLASSINFOpdf
   \usepackage[pdftex]{graphicx}
   \graphicspath{{./}}
   \DeclareGraphicsExtensions{.pdf,.jpeg,.png,.tiff}
\else
\fi
\usepackage{array}
\newcolumntype{P}[1]{>{\centering\arraybackslash}p{#1}}
\newcolumntype{M}[1]{>{\centering\arraybackslash}m{#1}}
\begin{document}
%
\title{Using Big Data to Enhance the Bosch Production Line Performance: A Kaggle Challenge}

\author{\IEEEauthorblockN{Ankita Mangal}
\IEEEauthorblockA{Materials Science and Engineering\\
Carnegie Mellon University\\
Pittsburgh, Pennsylvania 15289\\
Email: ankitam@andrew.cmu.edu}
\and
\IEEEauthorblockN{Nishant Kumar}
\IEEEauthorblockA{Data Scientist\\
Uber Technologies Inc.\\
San Francisco, USA\\
Email: nishant@uber.com}}

\maketitle

\begin{abstract}
This paper describes our approach to the Bosch production line performance challenge run by Kaggle.com. Maximizing the production yield is at the heart of the manufacturing industry. At the Bosch assembly line, data is recorded for products as they progress through each stage. Data science methods are applied to this huge data repository consisting records of tests and measurements made for each component along the assembly line to predict internal failures. We found that it is possible to train a model that predicts which parts are most likely to fail. Thus a smarter failure detection system can be built and the parts tagged likely to fail can be salvaged to decrease operating costs and increase the profit margins.
\end{abstract}


%
\IEEEpeerreviewmaketitle

\section{Introduction}
Smart manufacturing is being touted as the next industrial revolution \cite{Bryner2012}. With real time monitoring of manufacturing processes, to increase productivity and stay competitive, the use of data science methods is an obvious next step. For example, the defense manufacturing company Raytheon implemented the MES (manufacturing execution system) in its missile plant in Huntsville, ALA \cite{Hagerty2013}. This system collects and analyzes factory shop floor data, and was able to find out exactly how many times a screw needs to be turned in a critical component to be perfect. Big data can be used to predict equipment failure rates, streamline and optimize inventory management and prioritize processes. In 2012, Intel saved \$3 million in manufacturing costs through the use of predictive analytics to prioritize its silicon chip inspections \cite{EinatRonenKennethBurns2013a}. Smart manufacturing is the next big development following the well-established processes of Lean manufacturing and Six Sigma methodology. Following this trend, Bosch released its dataset consisting of anonymized records of measurements and tests made for each component along the assembly line, in a Kaggle competition \cite{Kaggle2016}, and challenged the Kaggle community to predict product part failures, thus enabling Bosch to bring quality products at lower costs to the end user. \hfill 

In this paper, we present our findings from the dataset. We explore the challenges faced due to the size of the dataset, the kind of data recorded, and machine learning algorithms that are suitable for these kind of problems. In section II, we will describe the dataset and the insights gained from exploring the data, section III provides an overview of the machine learning techniques used and section IV will go over our result and discussions.\hfill 

\section{Dataset}
Bosch has supplied a huge dataset (14.3 Gb) containing three types of feature data: numerical, categorical, date stamps and the labels indicating the part as good or bad. The training data has 1184687 samples and the learned model will be used to predict on a test dataset containing 1183748 samples. There are 968 numerical features, 2140 categorical features and 1156 date features. Hence, one of the biggest challenges of this dataset is to process these features into something meaningful so they can be used to make a predictive model.

\subsection{Categorical features}
The categorical data has 2140 features, but on further evaluation, we find that about 500 are multi value, 1490 single value and 150 are empty. The empty categorical features can be dropped as they contain no information. The single and multi-value categorical features can be converted to numerical by using the one hot encoding technique \cite{Kung2014}, where each class is represented by an integer. One hot encoding transforms a single variable with n observations and d distinct values into d binary variables with n observations each. Each observation indicates the presence (1) or absence (0) of the $d^{th}$ binary variable. Since the number of categorical features is very high, this makes it difficult for traditional machine learning algorithms to incorporate it by one hot encoding, because the feature space explodes into thousands of features thus becoming comparable to the total number of samples in the dataset.\hfill 

 \begin{figure}[!t]
\centering
\includegraphics[width=3.5in]{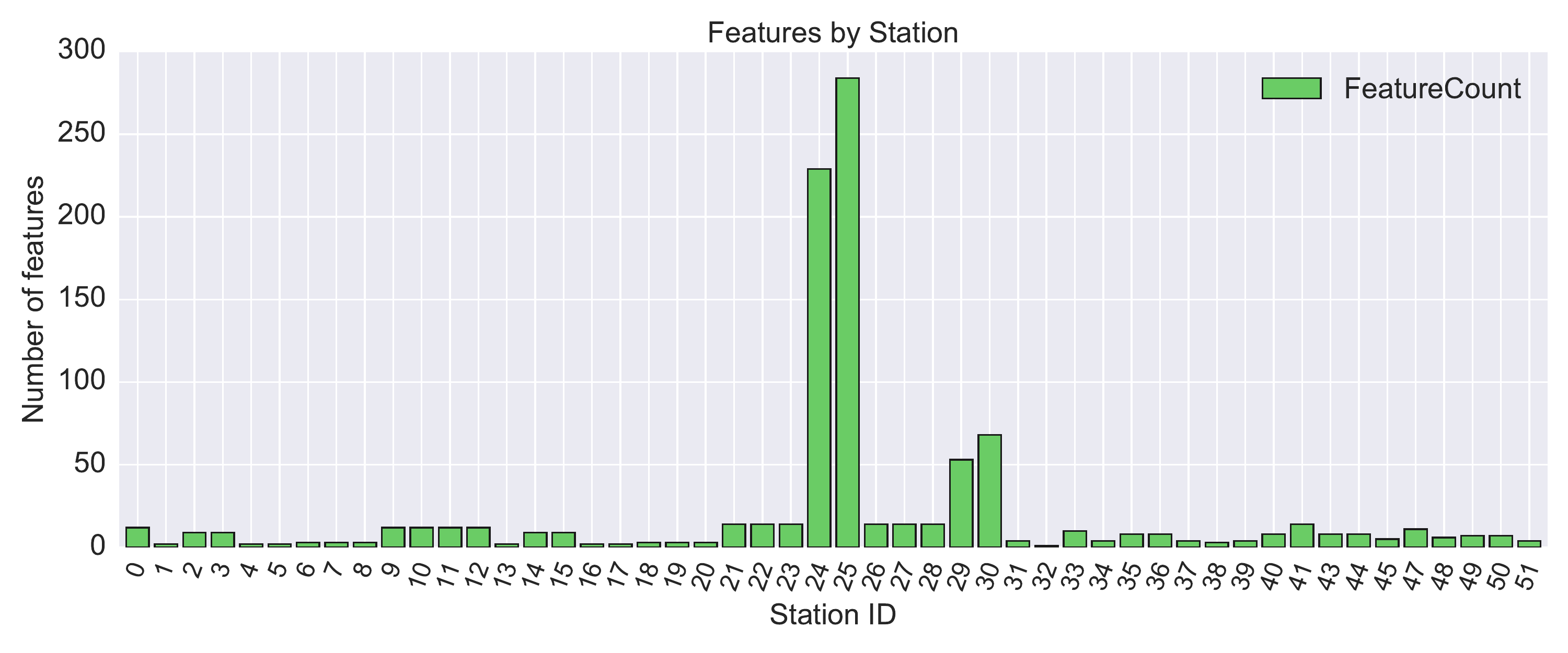}
\caption{Number of non-zero measurements in each station}
\label{fig. 1}
\end{figure} 

\subsection{Numerical features}
The numerical feature names contain information about the stations, production line and a test number combination. The value for that feature is the corresponding measurement. For example, a feature named L3\_S50\_F4243 for a component indicates that the part went through production line 3, station 50, and the feature value corresponds to a test number 4243. This way, each product coming out of the manufacturing line can be segregated according to the production flow. To prevent confusion, we will refer to these feature values as measurements. We observe there exist 51 stations distributed between 4 production lines. Counting the total number of non-zero measurements in each station (fig. 1), we see that station 24 and 25 have the most number of measurements ( $>200$), station 32 has only one measurement and remaining stations have about 20 measurements.\hfill 

To understand how parts are moving through the stations, a count of number of parts per station is plotted in fig. 2. We observe that each station has different number of parts passing through it, which could mean the existence of different classes of products, each going through a certain production path. Also, station 32 has very few parts going through it, which means it does not process many parts. This indicates that station 32 is some sort of a reprocessing or a post processing station. \hfill 
 
To explore if certain production lines or stations are correlated to higher error rates, we calculate the fraction of defective parts in each station and production line. 
Fig. 3 shows the percentage error between stations and we found that station 32 has the highest error rate. However, station 32 does not process a lot of products, hence its impact on the production yield is minimal. A total of 24,543 samples run through station 32, with a 4.7\% error rate, compared to an average error rate of about 0.6\% in the other stations. It only has one feature, L3\_S32\_F3850 which could mean that station 32 is a part of production line 3 and after reprocessing, all parts go through the same test indicated by F3850. Station 31 is associated with the lowest failure rate at 0.27\%. \hfill 

\begin{figure}[!htb]
\centering
\includegraphics[width=3.5in]{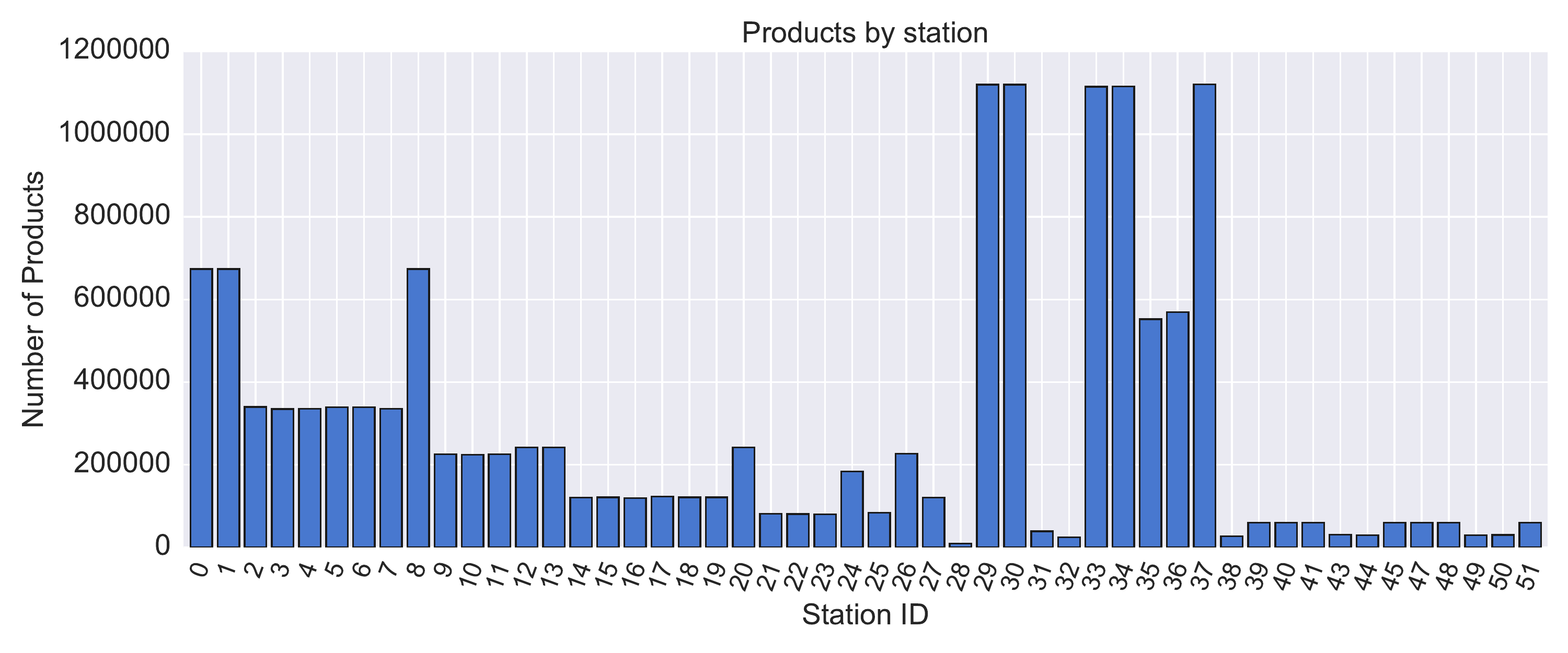}
\caption{Number of product parts passing through each station}
\label{fig. 2}
\end{figure}

\begin{figure}[!htb]
\centering
\includegraphics[width=3.5in]{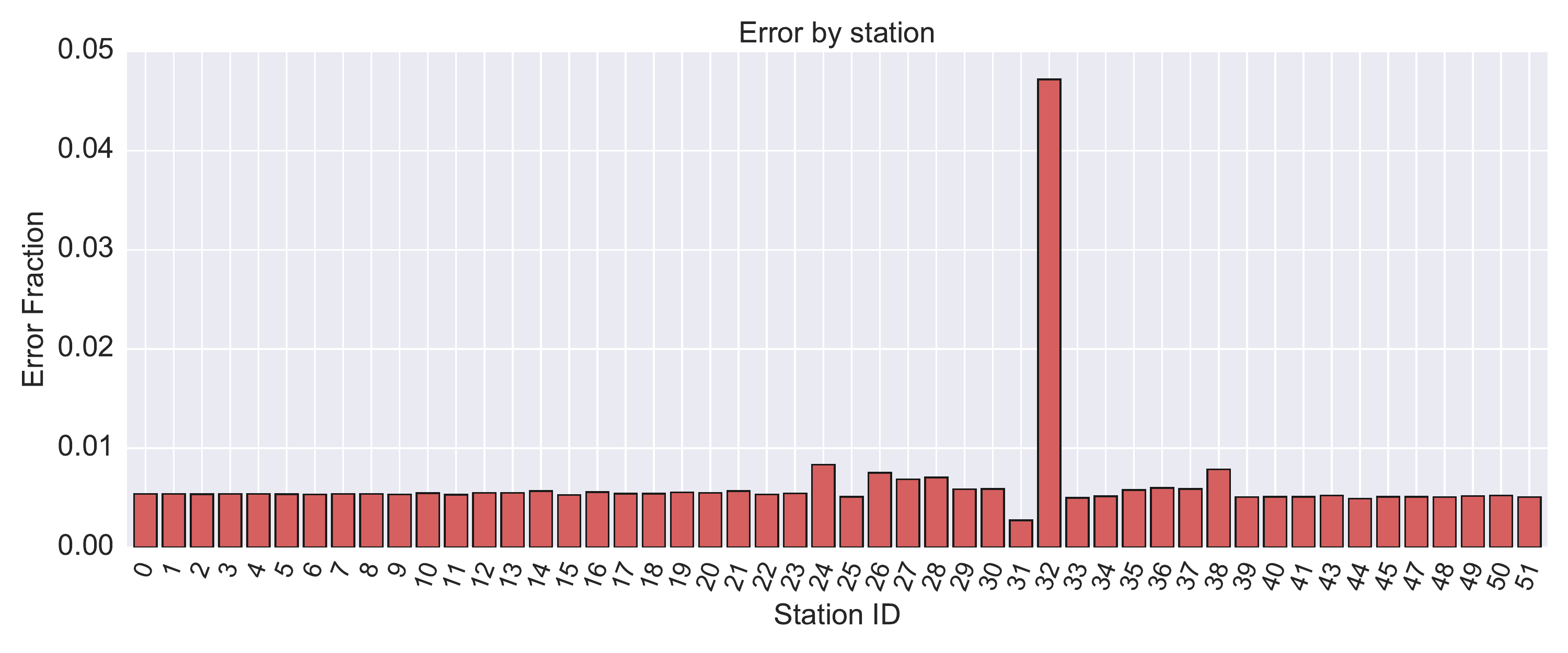}
\caption{Fraction of defective products in each station}
\label{fig. 3}
\end{figure}

\begin{figure}[!htb]
\centering
\includegraphics[width=3.5in]{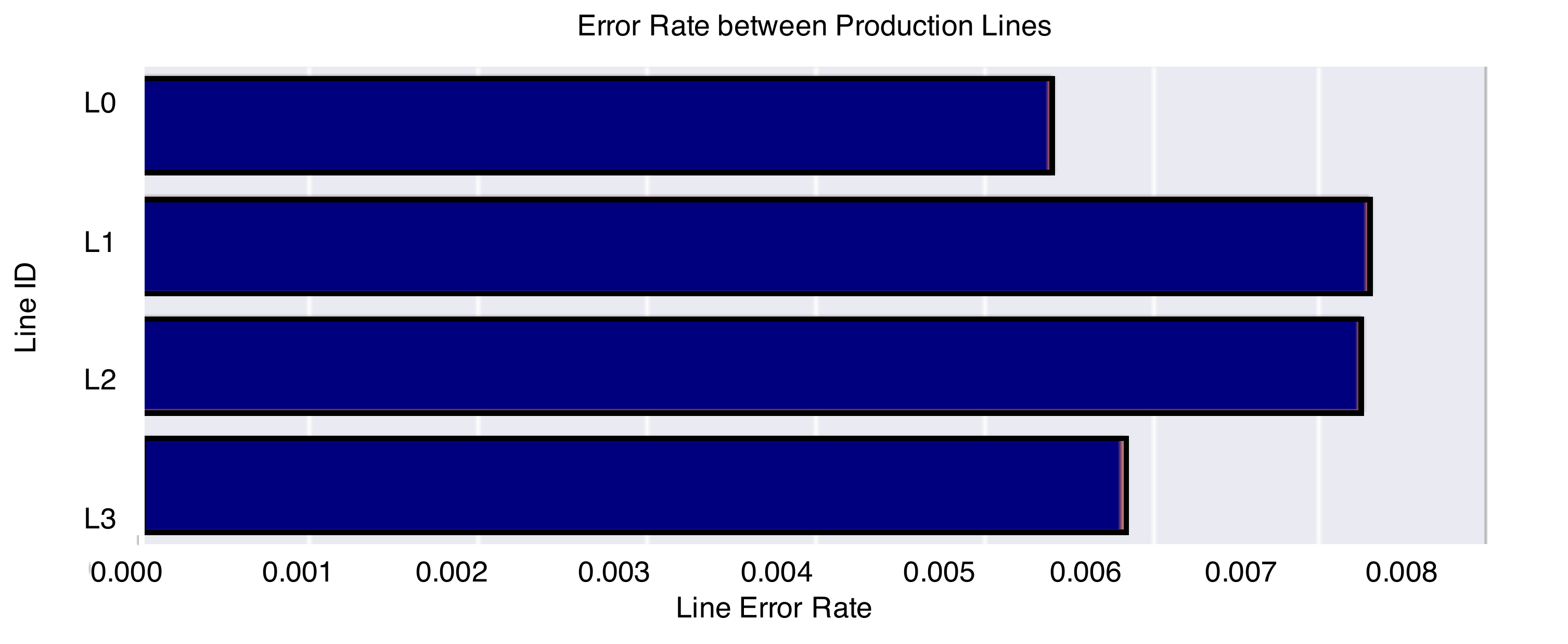}
\caption{Fractional error in each production line}
\label{fig. 4}
\end{figure}

\begin{figure}[!htb]
\centering
\includegraphics[width=3.5in]{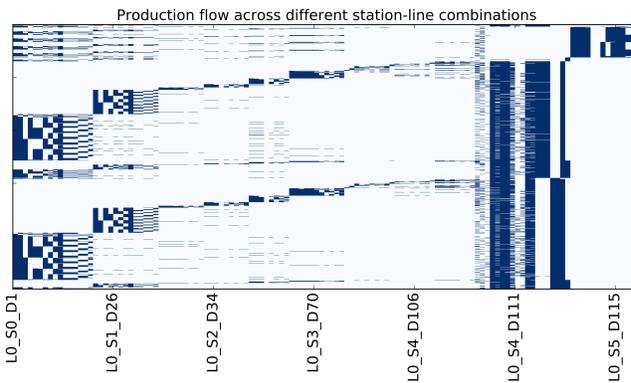}
\caption{Number of non-zero measurements made vs. the production line-station number-date combination}
\label{fig. 5}
\end{figure}

Fig. 4 explores error rates when compared to the different production lines and no line was found to have unusual error rate.\hfill 
 
To explore how the parts are moving through the production line, the samples were aggregated by line and station number and a production flow is revealed as can be seen in fig. 5. A total of 7148 unique flow paths were found. \hfill 

However, this means that there are several product categories in the dataset and in order to achieve the best predictions, models need to be fit to each of these product categories. Using the flow path data, these parts can be clustered into product families by grouping similar part frequency paths, however, that is outside the scope of this work. \hfill

Similar products would spend similar time in the production line. Using this intuition, we engineer a ``time\_diff'' feature indicating the total time spent for each part in the production line. To keep things simple, we will continue with fitting a general model to all the parts with the above mentioned caveat.\hfill

\subsection{Date features}
The date features names are labeled by production line, station id and date id. For example, L3\_S50\_D4242, would mean the product went through production line 3, station 50, and the feature value corresponds to date id 4242. There are a total of 1157 date features, with a lot of missing values. Same stations often have same date values. Fig. 6 shows the plot of the number of records vs. the date feature value. A clear periodic pattern can be observed in the data, with the date feature values lying between 0-1718, with a granularity of 0.01. \hfill

To understand the time periods corresponding to these numbers, the autocorrelation between the features is computed as a function of time lag between them (fig. 7). We observe that the largest peaks lie at a date feature value 16.75 ticks, and there are about 7 local maxima in between which should correspond to the days of a week. Thus 1 week is 16.75 date feature value units and data is recorded at a granularity of 6 minutes. Since the dataset corresponds to measurements taken over 1718.48 time units i.e. 102.6 weeks, this explains the variability in the dataset because the factory conditions can change with time.\hfill

\begin{figure}[!htb]
\centering
\includegraphics[width=3.4in]{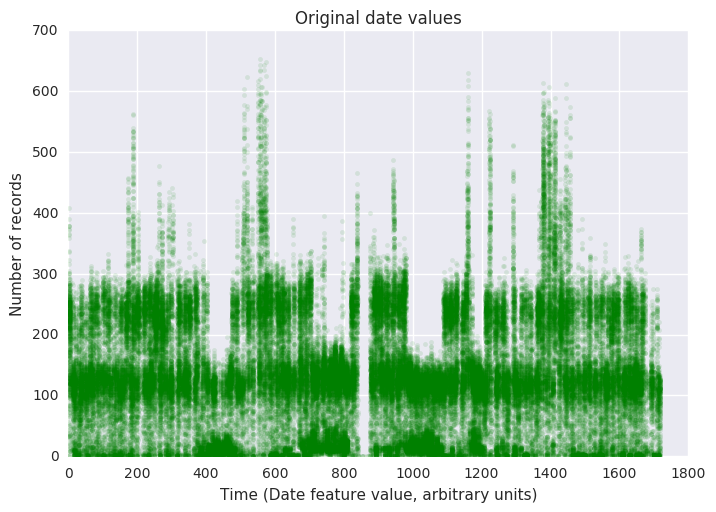}
\caption{Number of records made on the date feature value. A periodicity can be observed. The granularity is 0.01, which corresponds to ${\sim6}$ minutes in real time.}
\label{fig. 6}
\end{figure}

\begin{figure}[!htb]
\centering
\includegraphics[width=3.4in]{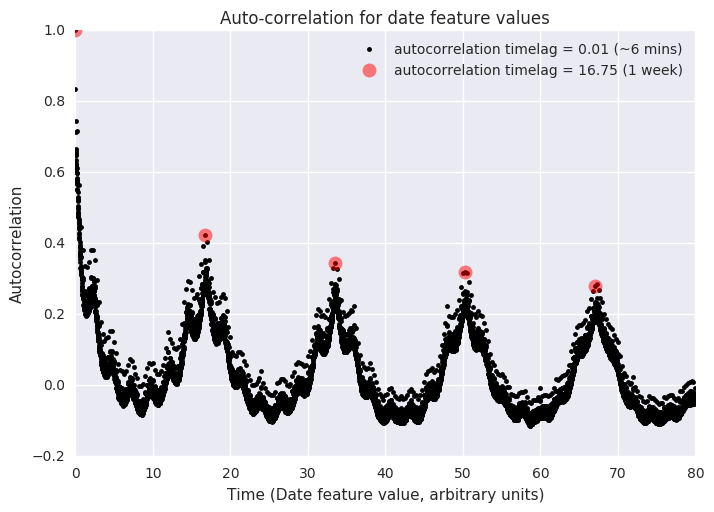}
\caption{Autocorrelation for the number of observations recorded on a day as a function of time lag between them. A periodicity is revealed, with bigger maxima (every 16.75 ticks) corresponding to weeks and smaller local maxima (every 2.39 ticks) corresponding to a single day.}
\label{fig. 7}
\end{figure}

\section{Machine Learning Techniques}
\subsection{Selecting a learning algorithm}
The large number of features in this dataset present a unique machine learning challenge due a comparable number of features and samples and memory constraints. There is a need to use methods for automated management of features and performance analysis for getting confidence estimates for predicted probabilities.\hfill

Online advertising is a multi-billion-dollar industry where predicting click through rates (CTR) is a central massive scale learning problem. The data in such problems is extremely sparse, the feature vectors might have billions of dimensions, but typically have a very tiny fraction of non-zero values. Drawing inspiration from algorithms used for such problems, i.e. using high dimensional categorical data for modeling rare events, we will use an ``online'' learning technique called Follow the Regularized Leader (FTRL) algorithm \cite{McMahan2013} for converting all the categorical features into a single numerical feature which is further used for stacking as a new feature\cite{Zenko2004} in the next step. The reduced feature space will now allow the use of conventional machine learning algorithms. Fig. 8 demonstrates this workflow.
\hfill

\subsection{Online learning and sparsity }
Online learning \cite{fontenla2013online} is a technique which is used when its computationally expensive to train the entire dataset in a batch or the algorithm needs to dynamically adapt to new patterns in the data. The training data becomes available sequentially and the model is updated each time a new data point becomes available. \hfill

FTRL (Follow the Regularized Leader) algorithm is a variant of Online learning which was especially used for our model. The FTRL algorithm allows use of L1 and L2 regularization \cite{Ng:2004:FSL:1015330.1015435} in its optimization which makes it very robust to overfitting. L1 (Lasso) helps with reducing a lot of features to zero whereas L2 (Ridge) helps in keeping all the coefficients of features low. To summarize the algorithm, it can be seen as a 3 step process:
\begin{itemize}
  \item construct a feature vector x, for every sample using the ``hashing trick'' \cite{Attenberg2009}
  \item compute a prediction using the current weight vector w and feature vector x, by taking their dot product
  \item update the model after calculating the gradient of the prediction with the label i.e $p - y$; p=predicted probability, y= label 1/0
\end{itemize}

The feature vector is typically constructed using ``the hashing trick'', which hashes each categorical feature into indices. The strings are modulo a very large number (D) to limit the memory usage. We used  D=$2^{28}$ corresponding to 2.15 Gb of RAM which provided a large enough margin for our hashing feature space \cite{Attenberg2009}.\hfill

The prediction is simply the dot product of the weight vector with the feature vector. This can be computed efficiently because the feature vector is sparse. (You multiply only the non-zero coefficients, instead of all D coefficients.). We also used a low global learning rate, along with high L1 and L2 regularization to prevent overfitting. \hfill

Log loss is the logarithm of the likelihood function for a Bernoulli random distribution. To maximize the likelihood of data, Log loss needs to be minimized. It can take a minimum value of 0 for perfect prediction and can range to infinity. The FTRL model based on categorical features yields a log-loss of 0.34 (${AUC\sim0.67}$) on a sample validation data.\hfill

\subsection{Stacking}
To handle the high cardinality of categorical features, FTRL algorithm was used to train a model using categorical features only and then stacking the probability predictions as a new column along with numerical and date features \cite{Zenko2004}. This serves as a feature reduction technique in which all the categorical features get represented by one numerical column.\hfill

Fig. 8 illustrates this technique where the training data is divided into two equal parts, Subset 1 \& 2. A separate model is trained on each training subset, and the trained model is used to generate a prediction probability on the other subset. These unbiased probabilities are stacked and used as a single feature in the next machine learning model. \hfill

Since we have two models trained on each subset, we have two predicted probabilities for the test set. Hence we simply take their arithmetic mean to obtain the test data probability column from categorical features. Thus we can obtain a new numerical feature for both train and test data.

\begin{figure}[!htb]
\centering
\includegraphics[width=3.5in]{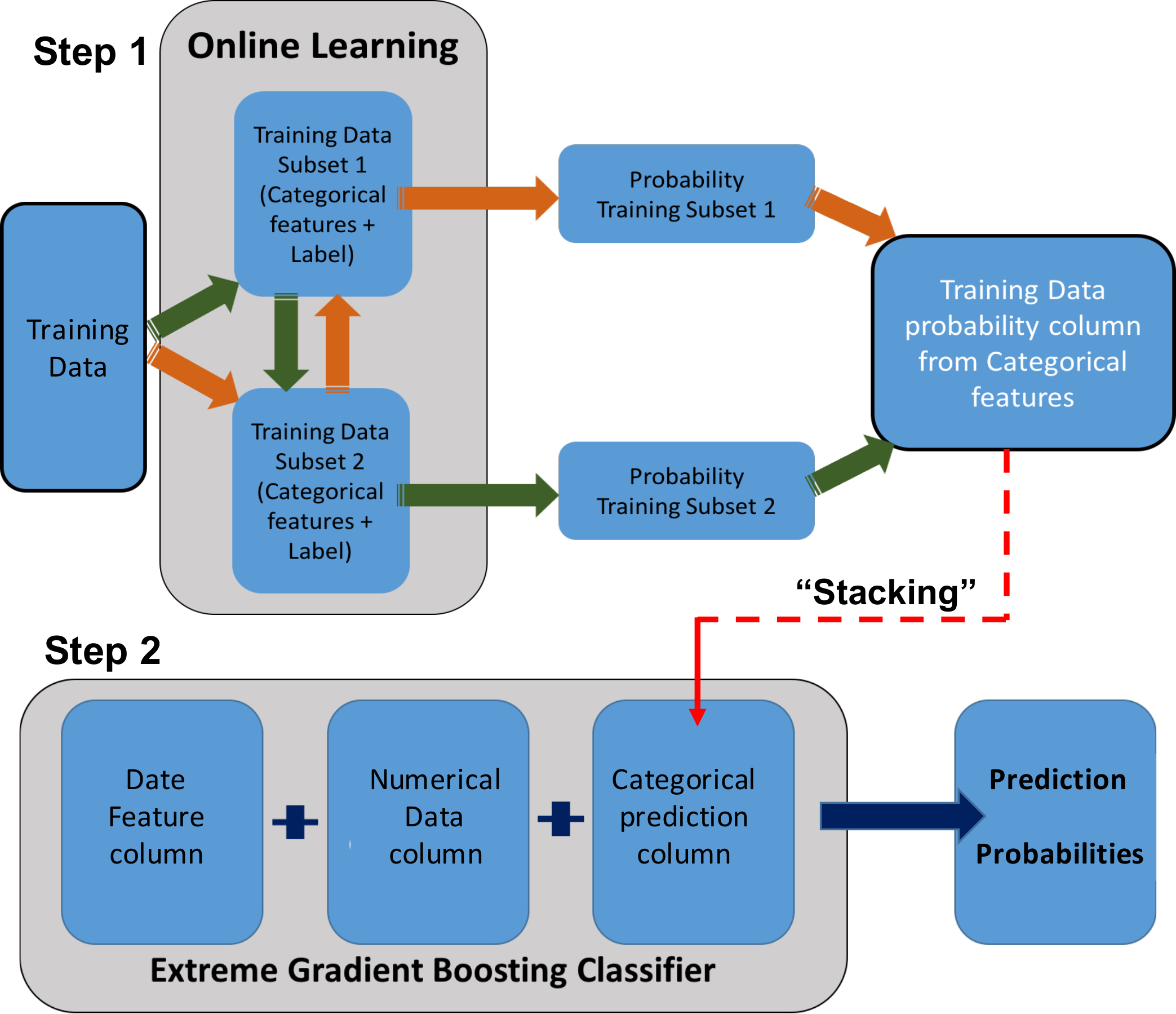}
\caption{Workflow for the techniques used. Step 1: Online learning where the training dataset is divided into two folds to obtain prediction probabilities in the other fold. These probabilities are stacked to form a new numerical column to be used in step 2: using XGBoost to predict the product failure probability using the transformed categorical, selected numerical and date features }
\label{fig. 8}
\end{figure}

\subsection{Extreme gradient boosting classifier (XGBoost)}
After converting all the categorical features into numerical type, the feature space dimension reduces to 968 original numerical features, 1 engineered feature,1 derived numerical feature and 1156 date features, making a total of 2126 features. This makes the use of conventional machine learning algorithms possible. We chose Extreme gradient boosting (A faster implementation of traditional Gradient boosted trees) classifier \cite{friedman2000additive}, \cite{friedman2001greedy},\cite{Chen2015}, \cite{Chen2016} because this is a non-linear algorithm which works very well with numerical features and requires less feature engineering and hyper-parameter tuning, which makes it simple to implement in this case.\hfill

Gradient Boosted Trees, a strong classification algorithm, is essentially an ensemble of multiple weak trees. The small depth trees are created on a sample of rows and features at each step and these trees are used to come up with a prediction. Using the current prediction, Log-loss cost function gradient is calculated with respect to the target and then the next round of trees are created to learn the gradient. Thus, it tries to minimize the error obtained till ${n_{th}}$ step. This method is prone to overfitting since it constantly involves fitting a model on the gradient. To prevent this, we optimize for the number of trees until the out of sample error starts increasing again. Some of the hyper-parameters we will be discussing for XGBoost are:

\begin{itemize}
  \item learning\_rate: Learning rate shrinks the contribution of each tree by learning\_rate. 
  \item n\_estimators: The number of boosting stages to perform. This parameter needs to be monitored on a neutral test set to prevent overfitting.
  \item max\_depth: Maximum depth of the individual regression estimators. The maximum depth limits the number of nodes in the tree. The best value depends on the interaction of the input variables.
  \item nthread: Number of cores to be used in parallel during model training. -1 stands for using all cores. We trained the model on an 8 core MacBook Pro 2015. 
  \item min\_child\_weight: Minimum sum of instance weight (Hessian) needed in a child. If the tree partition step results in a leaf node with the sum of instance weight less than min\_child\_weight, then the building process will give up further partitioning. In linear regression mode, this simply corresponds to minimum number of instances needed to be in each node. The larger this number, the more conservative the algorithm will be.
\end{itemize}
 
The second stage of learning is to select the most important features and then use the XGBoost (XGB) algorithm to predict the class probabilities, to which a threshold is applied to get the final class predictions.\hfill

\subsubsection{Feature Selection}
The feature importance measures are based on the number of times a variable is selected for splitting the component trees in XGBoost, weighted by the squared improvement to the model as a result of each split, and averaged over all trees. In other words, feature importance in tree ensemble models is given by how frequently a feature has appeared in the model trees.\hfill

Out of the remaining 2126 features, feature selection is performed using gradient boosting on a random subset of 100,000 samples of the training data to prevent sampling bias. The parameters used for this preliminary training are learning\_rate=0.1, max\_depth=3, min\_child\_weight=1, n\_estimators=100, nthread=-1.\hfill

This helps in selecting the top ${200/2126}$ features and then the final model is trained using the entire 1 million row training data with these 200 features. This two-step process helps in reducing run time and memory footprint of feature selection as well as final model training.

\subsubsection{Extreme Gradient Boosting Classifier Training}
Using these top 200 features, the whole training (${\sim1}$ million samples) dataset is used to train a new XGBoost  model \cite{Chen2015}, \cite{Chen2016}. The whole training data is divided randomly into three subsets of 33\% data each. Three separate XGB models with same hyper-parameters are trained on 67\% of the data and evaluated on the remaining 33\% data to reduce memory consumption and increase training speed. The predictions for each fold are stacked and are then used to come up with an unbiased estimate of the training error for the entire training set. This three-fold cross validation technique is used to fine-tune the hyper-parameters of the XGB model while optimizing the Area Under Curve (AUC). The final individual Area Under Curves for the three models are 0.719,  0.718, 0.718. After stacking all their predictions, the overall out-of-fold AUC for the entire training data becomes 0.718${\pm}$  0.001.\hfill

The parameters used for final XGB training are learning\_rate=0.01, max\_depth=7, min\_child\_weight=5, n\_estimators=100, nthread=-1. With the best tuned hyper-parameters in hand, the entire training dataset is used to train a final model. A similar approach of 3-fold cross-validation was tried with other classification algorithms, out of which XGB model performed the best as seen in table 1.\hfill

\begin{table}[!t]
\caption{Comparison of Different Algorithms on the Kaggle Bosch Training Dataset With 3-Fold Cross Validation}
\label{Table 1}
\centering
\begin{tabular}{|M{1.5in}|M{1.5in}|}
\hline
Classification Model Used & 3-fold Cross Validation Training AUC\\
\hline
Logistic Regression & ${0.614\pm0.004}$\\
\hline
Extra Trees Classifier & ${0.685 \pm0.003}$\\
\hline
Random Forest Classifier & ${0.709 \pm0.003}$\\
\hline
Extreme Gradient Boosting (XGB)  & ${0.718 \pm0.001}$\\
\hline
\end{tabular}
\newline \newline
\end{table}

\section{Results and Discussion}
The prediction probabilities for the entire training population are sorted descending, and then divided into deciles. For a random chance model, in order to cover half of the failed products, we need to validate 50\% of the entire products population. However, with our model, we can see from fig. 9 that if we target the top 2 decile failure probabilities from the model, we capture close to 50\% of our failure target population. This is much better than checking for defects at random.\hfill

We optimized our model for the best AUC which is a ranking measure of how the model is able to differentiate between the two classes of labels. However, even top 2 decile of the population can mean checking more than 200,000 parts for defects, which is financially and time-wise disadvantaged. Hence a better evaluation criterion would be the Matthews Correlation Coefficient (MCC) \cite{matthews1975comparison}. It is a measure of binary classification, and takes into account all elements of the confusion matrix (true and false positives and negatives).  Since product failure is a rare event, where less than 1\% of the population falls under the target class, it is suitable to use MCC as its a balanced measure of model performance. MCC ranges between -1 to +1. A coefficient of +1 represents perfect prediction, 0 is no better than random and -1 represents total disagreement between prediction and observation.\hfill

\subsection{Tuning for Matthews Correlation Coefficient}

\begin{figure}[!htb]
\centering
\includegraphics[width=3.5in]{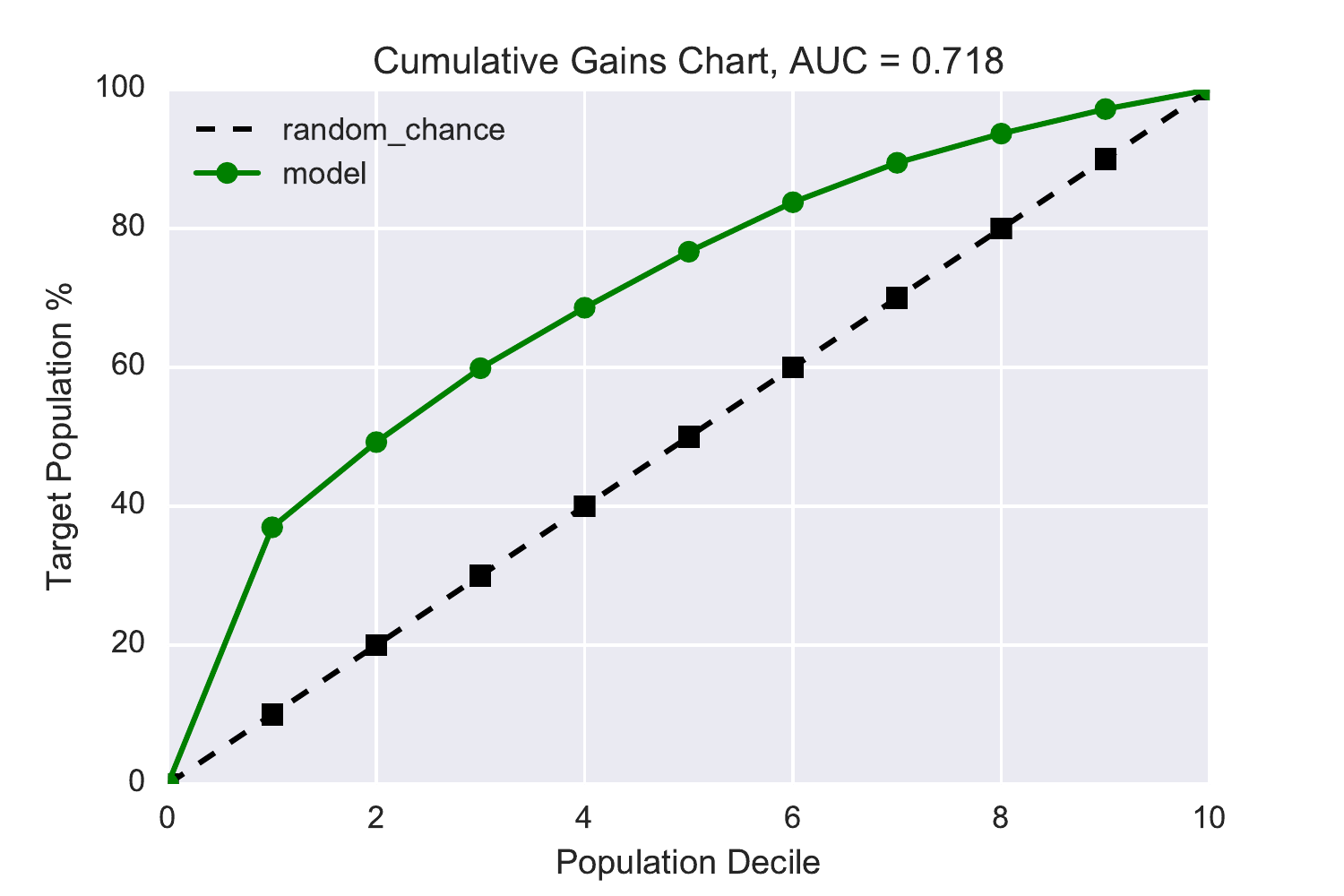}
\caption{The dotted line shows the population deciles to be evaluated for failure for random chance. Higher area under the solid curve signifies a better model prediction.}
\label{fig. 9}
\end{figure}

MCC is calculated for a score threshold that divides the population into the two classes. First the prediction scores are obtained for the training data using the 3-fold cross validation technique as described above. Then, to pick the best cutoff for Matthews coefficient, we iterate through all possible score cutoffs and calculate and plot the Matthews coefficient. From fig. 10, we can see that for training data, the best Matthews coefficient achieved was 0.227 at a probability threshold of 0.11. This MCC value corresponds to an AUC of 0.718.\hfill

This means that products having a prediction score greater than 0.11 can be tagged for post processing as they are most likely to fail. In the training dataset, only 3235 samples fall under this category and need to be re-evaluated as opposed to 2 deciles of population tagged by AUC. This results in saving time and resources as well as increased profit margins due to reduced product downgrading, increased salvage and higher production yields. 

\subsection{Understanding Feature Importances}
Some of the most important features in the final model are (in descending order):
\begin{itemize}
	\item L3\_S29\_D3316
	\item L3\_S29\_D3474
	\item L3\_S33\_D3856
	\item L3\_S34\_D3875
	\item L0\_S0\_F20
	\item L3\_S30\_F3774
	\item Probability derived from Categorical Features 
	\item time\_diff
\end{itemize}

\begin{figure}[!htb]
\centering
\includegraphics[width=3.5in]{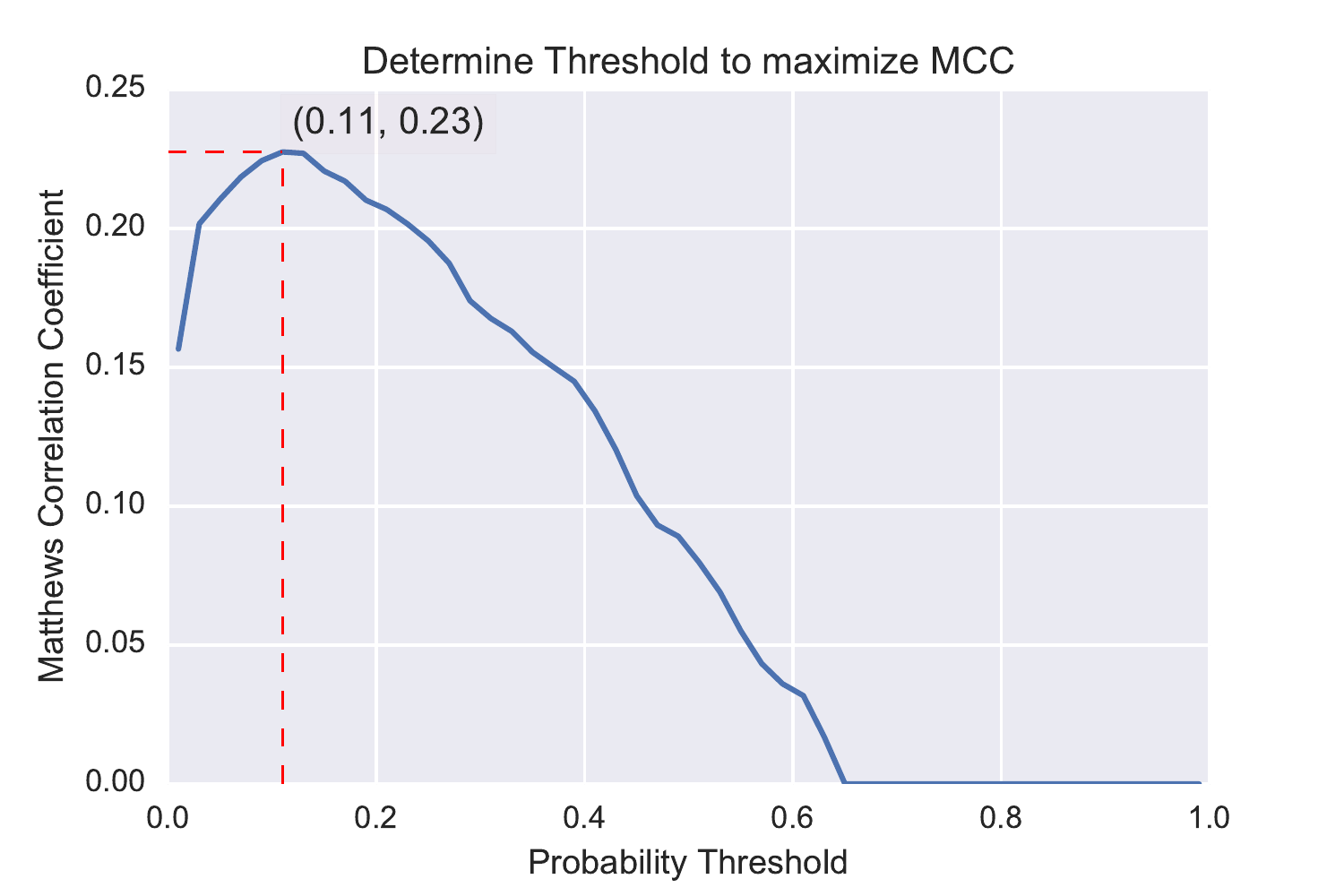}
\caption{Determining the probability threshold for maximizing the Matthews correlation coefficient for training dataset}
\label{fig. 10}
\end{figure}
If we look at the most important features in the final model, the categorical probability feature shows up among the most predictive features. We find that time\_diff is an important feature and has a positive correlation with failure rate. This is expected because a longer time in the production line is an indicator of the complexity of the part. We also see that production line 3 and date features are one of the most important features in the final model.  We strongly feel that there is a lot of scope for improving model through better processing of these date features which we omitted from our experiments here.

\subsection{Leaderboard Results}
On Kaggle public leaderboard (which is evaluated on a random 30\% sample of test set), the MCC score obtained using this model is 0.21514 and gives us a rank of ${\sim678/1283}$  teams playing as of ${3^{rd}}$ November 2016. Including additional time based features gives us a rank of ${\sim252/1283}$  with an MCC score of 0.40681 as of ${3^{rd}}$ November 2016 but they are outside the scope of this paper. Our team name is Hello World. 

\section{Conclusion and Future Work}
This dataset is sparse, with a lot of missing values, more than a million samples and a rare event with ${<1}$\% positive samples. This makes it computationally expensive to apply traditional machine learning techniques. The number of features are very high, and the feature space further explodes after one-hot encoding of the 2140 categorical features. Thus the dataset is very prone to overfitting due to the ``curse of dimensionality''. The presence of a majority of categorical features favors the use of an online learning model, which we have used here as a feature reduction technique. We obtained a training AUC of ${0.718\pm0.001}$, which is a good baseline model but has a lot of scope for improvement.\hfill

The overall runtime for Step 1 (online learning) was ${\sim1}$ hour and Step 2 (XGBoost) was ${\sim30}$ mins on an 8 core MacBook Pro 2015. \hfill

We found that there are 7158 unique flow paths along the production line and stations, which hints at the presence of several product categories. To achieve better predictions, the products need to be clustered into these categories and separate models should be made for each category. \hfill

We observed that there is a weekly periodicity to the number of observations recorded per day. This insight provides us an opportunity to improve the model by incorporating complex date based features taking into account the variability observed due to change in factory conditions over time.\hfill

\ifCLASSOPTIONcompsoc
  \section*{Acknowledgments}
\else
  \section*{Acknowledgment}
\fi

The authors would like to thank Kaggle user Belluga for posting their insights about the date-time data on the Kaggle competition forum \cite{Belluga2016}, and Kaggle user Dmitry Sergeev \cite{Sergeev} for insights about the production flow. This work was supported in part by the National Science Foundation under award number DMR-1307138.

\IEEEtriggeratref{8}
\IEEEtriggercmd{\enlargethispage{-4.3in}}


\bibliographystyle{IEEEtran}
\bibliography{Kaggle.bib}
%



\end{document}